\documentclass[twoside]{article}

\usepackage{microtype}
\usepackage{hyperref}

\usepackage{AVTcommands}
\usepackage{AVTenvironments}

\usepackage[accepted]{aistats2021}

\usepackage[authoryear]{AVTcitations}
\bibliography{bibliography.bib}

\usepackage{tikz}
\usepackage{subcaption}
\usepackage{enumerate}
\usepackage{booktabs}

\definecolor{plotblue}{RGB}{40,117,177}
\definecolor{plotred}{RGB}{251,28,33}

\newcommand{\modelparam}{\theta}
\newcommand{\nph}{n+\frac{1}{2}}
\newcommand{\npth}{n+\frac{3}{2}}

\renewcommand\UrlFont\scshape

\runningtitle{Learning Contact Dynamics using Physically Structured Neural Networks}
\runningauthor{Andreas Hochlehnert, Alexander Terenin, Steind\'{o}r S{\ae}mundsson, and Marc Peter Deisenroth}

\begin{document}

\twocolumn[

\aistatstitle{Learning Contact Dynamics using\\Physically Structured Neural Networks}

\aistatsauthor{Andreas Hochlehnert\And Alexander Terenin}
\aistatsaddress{Centre for Artificial Intelligence\\University College London\And Department of Mathematics\\Imperial College London}

\aistatsauthor{Steind\'{o}r S{\ae}mundsson\And Marc Peter Deisenroth}

\aistatsaddress{Department of Computing\\
Imperial College London\And Centre for Artificial Intelligence\\ University College London} ]

\begin{abstract}
Learning physically structured representations of dynamical systems that include contact between different objects is an important problem for learning-based approaches in robotics.
Black-box neural networks can learn to approximately represent discontinuous dynamics, but they typically require large quantities of data and often suffer from pathological behaviour when forecasting for longer time horizons. 
In this work, we use connections between deep neural networks and differential equations to design a family of deep network architectures for representing contact dynamics between objects.
We show that these networks can learn discontinuous contact events in a data-efficient manner from noisy observations in settings that are traditionally difficult for black-box approaches and recent physics inspired neural networks.
Our results indicate that an idealised form of touch feedback---which is heavily relied upon by biological systems---is a key component of making this learning problem tractable.
Together with the inductive biases introduced through the network architectures, our techniques enable accurate learning of contact dynamics from observations.
\end{abstract}

\section{Introduction}

Deep-learning-based models have accomplished remarkable achievements in a myriad of fields in recent years, ranging from image processing to text generation to reinforcement learning.
These models are increasingly being applied to model physical systems, in areas ranging from fluid dynamics \cite{kutz2017deep} to robotics \cite{viereck2018learning,lutter2020deep, alvarez2020dynode}.
Though neural networks are good at approximating general classes of functions, they often struggle to learn invariant properties of physical systems, such as the conservation of energy or momentum \cite{greydanus2019hamilton} and other qualitative properties.
A rapidly-growing line of work \cite{raissiDeepHiddenPhysics2018,, greydanus2019hamilton, lutter2020deep,cranmer2020lagrangian, saemundsson2020variational} has thus focused on how to introduce \emph{inductive biases} into these networks to enable them to learn more accurate models from less~data.

\begin{table}[b!]
\footnotesize\raggedright Code available at: \url{https://github.com/libeanim/contact-symplectic-integrator-network}.
\vspace*{6.19ex}
\end{table}

\begin{figure*}
    \centering
    \begin{subfigure}{0.49\textwidth}
        \centering
        \begin{tikzpicture}[scale=0.7]
            \draw [-, thick, dashed] (-1, 1) -- (5, -1);
            
            \draw [-latex, thick] (-1, 1) -- (-1, 0.3);
            \draw [thick, fill=white] (-1,1) circle (0.2);
            \fill[fill=black] (-2,0) -- (6,0) -- (6,-0.1) -- (-2,-0.1);
            \node at (-1, -2) {$t_0$};
            \draw [-latex] (-0.5, -2) -- (1, -2);
            
            \draw [-latex, thick] (1.5, 0.2) -- (1.5, -0.5);
            \draw [thick, dashed, fill=white] (1.5,0.2) circle (0.2);
            \node at (1.5, -2) {$t_c$};
            \draw [-latex] (2, -2) -- (4.5, -2);
            
            \draw [-latex, thick] (5, -1) -- (5, -1.6);
            \draw [thick, fill=white] (5,-1) circle (0.2);
            \node at (5, -2) {$t_1$};
        \end{tikzpicture}
        \caption{Find the contact time $t_c$.}
    \end{subfigure}
    \begin{subfigure}{0.49\textwidth}
        \centering
        \begin{tikzpicture}[scale=0.7]
            \draw [-latex, thick] (-1, 1) -- (-1, 0.3) node[left]{};
            \draw [thick, fill=white] (-1,1) circle (0.2);
            \fill[fill=black] (-2,0) -- (0,0) -- (0,-0.1) -- (-2,-0.1);
            \node at (-1, -1.2) {$t_0$};
            \draw [-latex] (-0.5, -1.2) -- (1.5, -1.2);
            
            \draw [-latex, thick] (2, 0.2) -- (2, -0.5) node[left]{};
            \draw [thick, fill=white] (2,0.2) circle (0.2);
            \fill[fill=black] (1,0) -- (3,0) -- (3,-0.1) -- (1,-0.1);
            \node at (2, -1.2) {$t_c^-$};
            \draw [-latex] (2.5, -1.2) -- (4.5, -1.2);
            
            \draw [-latex, thick] (5, 0.2) -- (5, 0.9)node[left]{};
            \draw [thick, fill=white] (5, 0.2) circle (0.2);
            \fill[fill=black] (4,0) -- (6,0) -- (6,-0.1) -- (4,-0.1);
            \node at (5, -1.2) {$t_c^+$};
            \draw [-latex] (5.5, -1.2) -- (7.5, -1.2);
            
            \draw [-latex, thick] (8, 1) -- (8, 1.7)node[left]{};
            \draw [thick, fill=white] (8, 1) circle (0.2);
            \fill[fill=black] (7,0) -- (9,0) -- (9,-0.1) -- (7,-0.1);
            \node at (8, -1.2) {$t_1$};
        \end{tikzpicture}
        \caption{Calculate true trajectory.}
    \end{subfigure}
    \caption{Example integration scheme for contact dynamics that enforces constraints exactly, in the context of a bouncing ball. Initially, the ball is time-stepped until a contact with the floor is detected through interpenetration at time $t_1$. Then, the trajectory is (a) linearly interpolated to find the \emph{contact time} $t_c$ where contact occurs between the ball and floor. Finally, the contact state at time $t_c$ is calculated, a transfer of momentum between $t_c^-$ and $t_c^+$ is performed, and the ball is time-stepped as usual to time $t_1$.}
    \label{fig:contact-dynamics}
\end{figure*}
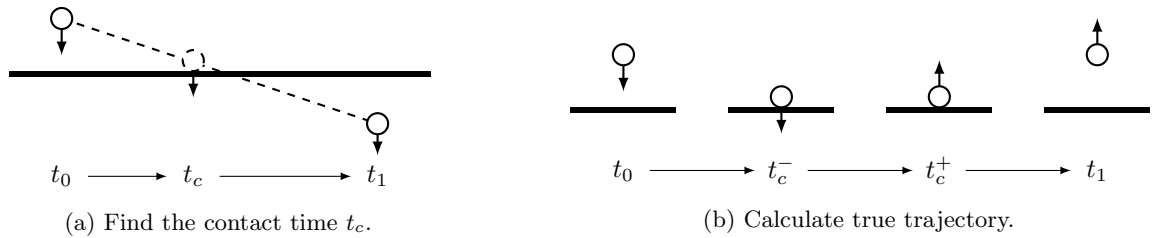

One particular kind of physical phenomenon of great interest to areas such as robotics is \emph{contact dynamics}, which describes how collisions between different objects affect the evolution of the system. 
For example, many of the basic actions that could be relevant for a robot, e.g. walking, jumping or grasping, involve discontinuous contact events.
Accurately modelling these dynamics, and related downstream phenomena, such as friction, is a crucial step towards enabling the creation of robots that can learn to interact with unknown objects in the real world. Additionally, stronger inductive biases can improve the data efficiency of such models \cite{deisenrothGaussianProcessesDataEfficient2015,greydanus2019hamilton,lutter2020deep,cranmer2020lagrangian, saemundsson2020variational}.
Due to the presence of non-linear and non-smooth behaviour, accurate multi-body contact dynamics are widely considered notoriously difficult to compute \cite{cirak2005decomposition}, due to challenges with numerically enforcing non-interpenetration constraints and other issues.
A robust and mature literature in physics and numerical analysis for handling these issues has developed in recent decades \cite{jean1999non, cirak2005decomposition, leyendecker2008variational}.

By bringing these ideas together with the rapidly-developing literature on neural ordinary differential equations \cite{E2017,Haber2017,chen2018neural,Ruthotto2018} and physically-inspired neural networks \cite{raissiDeepHiddenPhysics2018,, greydanus2019hamilton, lutter2020deep, saemundsson2020variational}, we study the problem of \emph{learning} unknown contact dynamics from data.
Our work is based on the approach of \textcite{saemundsson2020variational}, which derives neural network architectures by discretising the variational principle underlying the physical equations of motion under study.
Specifically, we propose specialised neural network architectures for modelling contact dynamics. 
These architectures combine discretisation schemes designed for contact dynamics with flexible parameterised networks for efficiently and accurately learning system behaviour.
We study these networks under different scenarios, develop schemes to ensure accurate learning of dynamics, and demonstrate empirically that the addition of an \emph{idealised touch feedback sensor}---rarely explicitly considered in deep learning, but widely utilised by biological systems---significantly improves model performance.

This suggests that the precise details of what hardware sensors the robot has available and how the learning problem is formulated to utilise those sensors, are both likely to have a significant impact on machine learning performance and should be studied further.
This includes understanding the effect of different forms of touch feedback, such as observed tactile sensors, and inferred feedback using proprioceptive sensors \cite{rotella2018unsupervised, ortenzi2016kinematics}.
Our work provides a starting point for addressing these questions within a physically structured deep learning framework.

\section{Contact Dynamics}
\label{sec:cd}

\begin{figure*}
\centering
\includegraphics{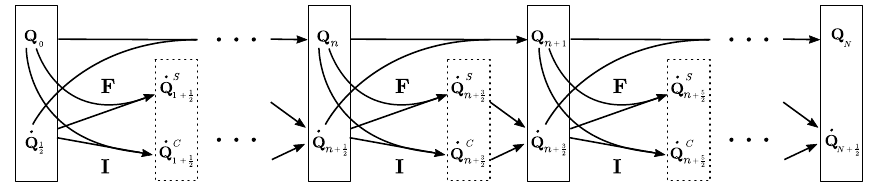}
\caption{A CD-Lagrange network. 
Here, we begin from initial states $(\m{Q}_0,\m{\dot Q}_{\frac{1}{2}})$. 
We calculate the next position $\m{Q}_1$, and proceed to calculate the next velocity $\m{\dot Q}_{1 + \frac{1}{2}} = \m{\dot Q}_{1 + \frac{1}{2}}^S + \m{\dot Q}_{1 + \frac{1}{2}}^C$ as a sum of smooth and contact terms.
These terms are in turn calculated using the conservative forces $\m{F}$ and the impulse $\m{I}$, which are calculated from the parameterised Lagrangian, whose potential energy is given by a fully connected network.
}
\label{fig:cdl-schema}
\end{figure*}

Here we briefly review the mathematical formulation of contact dynamics.
The state of a physical system is defined through position-velocity pairs $(\v{q},\v{\dot{q}})$.
From analytical mechanics, the trajectories of a physical dynamical system are assumed to be a stationary point of the action functional
\[
S(\v{q}, \v{\dot{q}}) = \int_{t_0}^{t_1} L(\v{q}_t,\v{\dot{q}}_t) \d t,
\label{eq:action integral}
\]
where $L$ is the Lagrangian---typically, the difference between kinetic and potential energy.
By the d'Alembert--Lagrange principle, at instances where there is no contact, these trajectories follow the \emph{Euler-Lagrange equations}
\[
\pd{L}{\v{q}} - \od{}{t} \pd{L}{\v{\dot{q}}} = \v{0}
.
\label{eq:euler-lagrange}
\]
These equations possess a number of important physical properties, such as conservation of energy, momentum, and phase volume.
A long line of work in numerical analysis has built efficient numerical integration schemes for these equations by maintaining these properties under discretisation \cite{marsden2001discrete}.
We are particularly interested in \emph{symplectic integrators}, which conserve phase volume exactly and conserve energy and momentum to a high order of accuracy \cite{sanz1992symplectic}.

At time points, where there is contact, the variational principle is augmented with contact constraints that depend on the precise physical setting.
These constraints ensure that states which are assumed impossible, such as those with interpenetration or deformation, cannot occur.
At these time points, the d'Alembert--Lagrange principle does not apply, and the Euler--Lagrange equations \eqref{eq:euler-lagrange} do not hold.
To handle this, one splits the action integral in equation \eqref{eq:action integral} into contact-free intervals with non-smooth contact-driven transitions in between, which are analysed separately.

Applying the variational principle at the contact time points yields \emph{transfer of momentum} equations, usually resulting from Newton's restitution law and the law of conservation of momentum \cite{fetecau2003nonsmooth, halliday2013fundamentals}.
These equations relate the state directly before contact to the state directly after contact.
To calculate this, one isolates the components of momentum, which are normal to the contact surface, and transfers them appropriately.

Implementing contact dynamics numerically yields a number of issues, which depend on the regime employed.
For example, certain regimes involve calculating the precise time when contacts occur, which might happen between integration time steps. 
In such cases, transfer of momentum is applied at the contact time point.
This yields good numerical accuracy, but often requires the solution of an optimisation problem to determine the precise contact time \cite{fetecau2003nonsmooth}.
Other regimes might instead relax the dynamics to allow object interpenetration in order to avoid expensive fixed-point iterations \cite{fekak2017new}.
This entails carefully considering how to handle interpenetration to ensure accuracy.
Figure \ref{fig:contact-dynamics} illustrates a sample numerical trajectory of a bouncing ball, including numerical transfer of momentum at a contact time point.

\section{Physically Structured Networks for Contact Dynamics}

To define neural network models for contact dynamics, we begin with the perspective of neural ODEs \cite{E2017, Haber2017, chen2018neural}, which will be helpful for this purpose.
Here, we specify a system of ODEs driven by a single-layer neural network, and discretise it to obtain a deep or recurrent neural network architecture.
In the classical case, an Euler discretisation yields a deep residual network, where the depth is given by the number of discretisation steps.

Building on these ideas, a number of recent works have proposed replacing the black-box system of ODEs with other systems that are more structured \cite{raissiDeepHiddenPhysics2018, greydanus2019hamilton,lutter2020deep,cranmer2020lagrangian,saemundsson2020variational}.
By applying structure-preserving discretisation schemes to these ODEs, one obtains neural network architectures with built-in inductive biases that improve generalisation and allow the networks to learn with less data.

In physical settings, a number of recent works have paired structured equations from mechanics, such as the Euler--Lagrange equations or Hamilton's equations, with structure-preserving integrators, such as the St\"{o}rmer--Verlet method, giving rise to \emph{physically structured neural networks} \cite{greydanus2019hamilton,lutter2020deep,cranmer2020lagrangian,saemundsson2020variational}.
These networks use the mathematical structure of classical mechanics as an inductive bias, ensuring the network mirrors important qualitative physical properties, such as conservation of momentum or conservation of energy.
These inductive biases have been shown to improve data efficiency and facilitate accurate long-term forecasting \cite{raissiDeepHiddenPhysics2018, greydanus2019hamilton,lutter2020deep,cranmer2020lagrangian,saemundsson2020variational}.

\subsection{Central-Difference Lagrange Networks}

We now employ these techniques to design neural network architectures specifically suited for modelling contact dynamics.
There are three main properties present in the true physics that we aim to encode into the neural network architecture.

\1[(a)] The network should be expressive enough to model a general Newtonian rigid body system.
\2 The network should be constrained to exclude non-physical dynamics that black-box networks allow, so that it learns in a data-efficient manner.
\3 The network should be able to handle periods without contact separately from contact events, as said phenomena differ in character.
\0 
Following \textcite{saemundsson2020variational}, one can construct a network satisfying the first two properties by applying a \emph{symplectic integrator} to the Euler--Lagrange equations induced by a free-form Lagrangian parameterised by a fully connected network, which are interpreted as a structured neural ODE.
We extend these constructions to explicitly handle contact events.
We begin by defining a parameterised Lagrangian $L_{\modelparam}$ for a free-form Newtonian potential system, given by
\[
L_{\modelparam}(\v{q}, \v{\dot{q}}) = \v{\dot{q}}^T \m{M} \v{\dot{q}} - V_{\modelparam}(\v{q}),
\]
where $\m{M}$ is a symmetric positive semi-definite position-independent inertia matrix, and $V_\theta$ is modelled by a fully connected neural network.

To obtain a network architecture for contact dynamics, we apply the symplectic \emph{Central-Difference Lagrange integrator} of \textcite{fekak2017new}, modified for rigid body impacts by \textcite{di2019benchmark}. 
We work with a simplified variant designed for modelling rigid non-deformable bodies---see Appendix \ref{apdx:cdl}, as well as \textcite{fekak2017new,di2019benchmark} for the general case.
This construction yields a recurrent network architecture specialised for modelling contact dynamics, which we now showcase. A schematic overview can be found in Figure \ref{fig:cdl-schema}.

CD-Lagrange uses separate time grids for the position $\v{q}_n$ and velocity $\v{\dot{q}}_{\nph}$, at times $t_n$ and $t_{\nph}$. We combine the position of all bodies in the matrix
\[
\m{Q}_n = \left(\v{q}_n^1,\ldots,\v{q}_n^K\right),
\]
where $K$ is the total number of bodies in the system.
Given a position-velocity pair $(\m{Q}_n, \m{\dot Q}_{n+\frac{1}{2}})$, CD-Lagrange calculates the next position using the midpoint velocity, given by
\[
\m{Q}_{n+1} = \m{Q}_n + \dfrac{h}{2} \m{\dot Q}_{n+\frac{1}{2}},
\]
where $h$ is the size of the time step. 
At the next time step, the change in velocity is calculated as a sum of changes due to smooth dynamics and due to contact, yielding
\[
\m{\dot Q}_{\npth} &= \m{\dot Q}_{\npth}^S + \m{\dot Q}_{\npth}^C,
\\
\m{\dot Q}_{\npth}^S &= \m{\dot Q}_{\nph} + h \m{M}^{-1} \m{F}(\m{Q}_{n+1}, \m{\dot Q}_{\nph} ), \label{eq:cdl-smooth}
\\
\m{\dot Q}_{\npth}^C &= \m{M}^{-1}\m{I}(\m{Q}_{n+1}, \m{\dot Q}_{\nph}), \label{eq:cdl-non-smooth}
\]
where $\m{F}$ is the conservative force and $\m{I}$ the impulse that occurs during contact, both defined below.
The conservative forces are calculated from the parameterised Lagrangian as
\[\label{eq:cdl-conservative}
\m{F}(\m{Q}_{n+1}, \m{\dot Q}_{\nph}) &= -\pd{L_{\modelparam}(\m{Q}_{n+1}, \m{\dot Q}_{\nph})}{\m{Q}_{n+1}}
\\
&= -\pd{V_{\modelparam}(\m{Q}_{n+1})}{\m{Q}_{n+1}},
\]
which, in our setting, are the conservative forces arising from the potential $V_{\modelparam}(\m{Q}_{n+1})$.
Rigid-body impacts are handled by Newton's restitution law \cite{fekak2017new, di2019benchmark}
\[
\m{U}_{\npth} = -e \m{U}_{\nph},\label{eq:restitution-law}
\]
where $\m{U}_{\nph} = \m{\dot Q}_{\nph} \m{L}^T_{n+1}$ with $\m{L}$ defined below, and $e\in [0,1]$ is the elasticity parameter with $e=1$ defining elastic impacts with no dissipation. 
Furthermore, for rigid body-body impacts, the law of conservation of momentum 
\[
\sum_{k=1}^K m^{k} \v {\dot q}^{k}_{\npth} = \sum_{k=1}^K m^k \v{\dot q}^k_{\nph}
\]
needs to be considered in order to uniquely resolve collision events.
Define the projection operator 
\[
\m{L} = \begin{bmatrix}
\v{n}_1
&
\hdots
&
\v{n}_k
&
\hdots
&
\v{n}_K
\end{bmatrix},
\]
which contains the normal vectors of the surface each body it is in contact with.
For each body $k$, the corresponding impulse $\v{I}^k$ is given by
\[
\v I_{n+1}^k = \begin{cases}
    \v L^k_{n+1} \lambda^k_{\npth} & \text{if } c^k_{n+1} = 1 \\
    \v 0 & \text{otherwise}  
\end{cases},
\]
where $c^k_{n+1}$ is a discrete contact signal for body $k$ and
\[ \label{eq:r_k}
    &\lambda^k_{\npth} = \left[\m H_{n+1} \left(e \m{\dot Q}_{\nph} + \m{\dot Q}^S_{\npth}\right)\m L_{n+1}^T \right]_{kk}
\]
represents the impulse acing on body $k$. The operator $\m{H}$ is defined as
\[
\left[\m{H}_{n+1}\right]_{i} = \left[ \m{A}_{n+1} \m{M}^{-1} \m{A}^T_{n+1} \right]_i^{-1},
\]
and ensures that the mass ratios between the bodies in contact, resulting from the law of conservation of momentum, are applied to the correct bodies.
Here, the operator $\m A$ selects the components of the bodies that are in contact with each other, and is given by
\[
[\m A_n]_{ij} =
\begin{cases}
-1 & \text{if  $i = j$ and } c_n^i = 1\\
1  & \text{if body $i$ and $j$ are in contact}\\
   & \text{(implicitly $c_n^i = c_n^j = 1$)}\\
0  & \text{otherwise.}
\end{cases}
\]
In total, these expressions define the general \emph{CD-Lagrange network}.
Further details, including derivation of these expressions, are provided in Appendix \ref{apdx:cdl}.

\paragraph{Idealised touch feedback.}

When using a CD-Lagrange network to predict future system states, one needs to determine whether or not objects are in contact at each time step in order to calculate whether or not the impulse component of the network comes into play.
To do so, we introduce an additional \emph{contact network} $\v{\hat{c}}_\theta$ that learns to predict a discrete contact signal $\v{c} \in \{0,1\}^K$ defined by
\[\label{eq:touch-sensor}
c^k_n = 
\begin{cases}
1 & \text{if contact for body } k
\\
0 & \text{otherwise}
\end{cases}
\]
at time step $n$. 
We consider two regimes: one in which $\v{c}_n$ is unobserved, and another where it is fully observed at training time. 
The latter case can be conceptually thought of as the addition of an \emph{idealised touch feedback} sensor to the system, which determines whether or not contact is present.
We explore this difference and its effect on performance in the sequel.

\begin{figure*}[t!]
    \centering
    \includegraphics[scale=0.45]{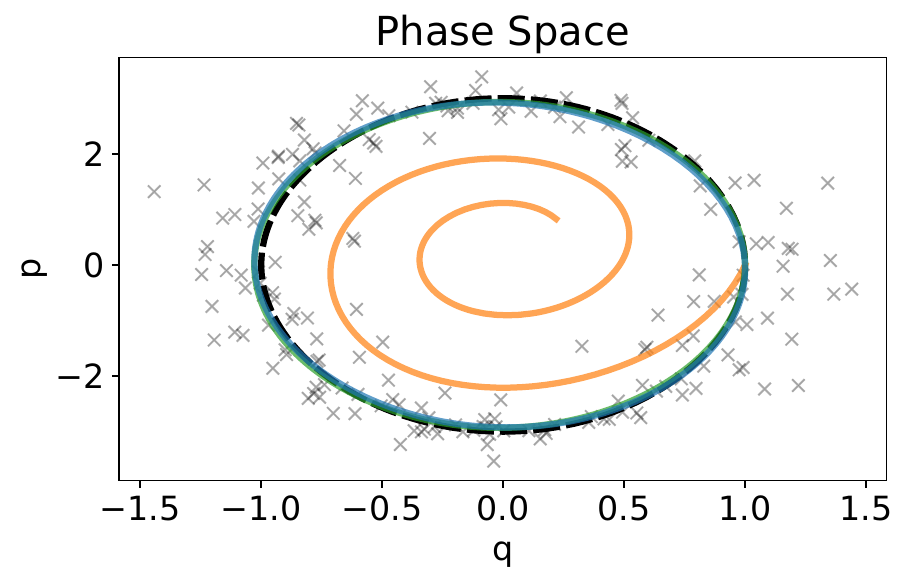}
    \includegraphics[scale=0.45]{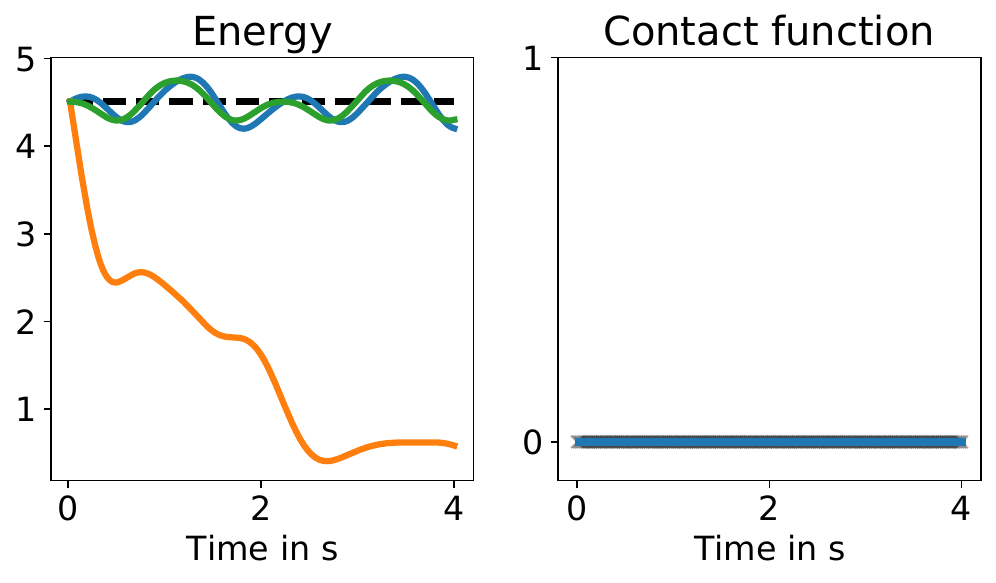}
    \includegraphics[scale=0.5]{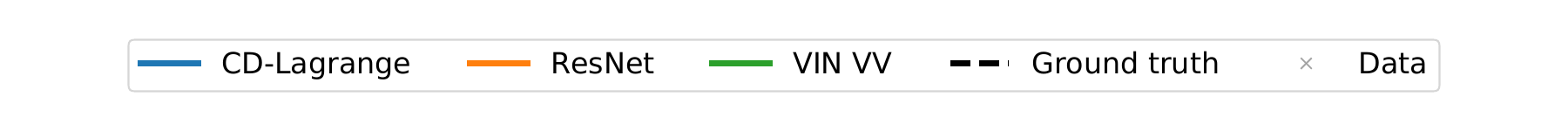}
    \caption{Learning the equations of motion of an ideal pendulum, which has no contacts. Given a set of initial conditions, we forecast a path in phase space and predict against ground truth. We display ground truth, training data, and model predictions, and energy over time for each of the models. For the CD-Lagrange network, we display the learned contact function, which is approximately zero everywhere. Root mean squared errors for each network are given as follows: \textsc{CD-Lagrange RMSE: 0.538, ResNet RMSE: 1.156, VIN VV RMSE: 0.509.}}
    \label{fig:pendulum}
\end{figure*}

\subsection{Learning from noisy observations}

Given a dataset of $J$ observed trajectories $(\v{y}_j, \v{c}_j)$, $j=1,\dotsc, J$, of length $N_j$ each with step size $h$ and $\v{y}_j = (\m{Q}_j,\m{\dot Q}_j)$. 
We train the network ${\hat{y}}_{\modelparam,h}$ by minimising the mean squared error loss between the (noisy) state observations and predicted states with respect to the parameters of the potential network $V_{\modelparam}$. 
We also jointly train the contact network $\hat{c}_{\modelparam}$ by minimising cross-entropy loss between its output and the contact signal $c$. For a single trajectory, this is given by
\[
\c{L}_{T}(\v{y}_j, \modelparam) = \frac{1}{N_jD} \sum_{n=1}^{N_j} \norm{\v{y}_n - {\hat{y}}_{\modelparam,h}(\v{y}_0, \v{c}_0)}^2
.
\]
The corresponding contact network loss is given by
\[
\c{L}_{C}(\v{c}_j,\v{y}_j,\modelparam) =-\frac{1}{N_jK} &\sum_{n=1}^{N_j}\sum_{k=1}^K \Big( c_n^k\log \hat{c}_{\modelparam}^k(\v{y}_n) 
\\
& + (1-c_n^k)\log(1 - \hat{c}_{\modelparam}^k(\v{y}_n)) \Big),\nonumber
\]
where the activation of the output of the contact network $\v{\hat{c}}_{\modelparam}$ is the sigmoid (logistic) function, and $\hat{c}_{\modelparam}^k(\v{y}_n)$ is the contact network's prediction for body $k$ at time step $n$. 
We further add a regularisation term $\c{L}_{R}(\modelparam)$ on the parameters of the model, which is described in the sequel.
In the full dataset, the observed trajectories might contain many steps or have unequal lengths $N_j$.
To avoid vanishing gradients, instead of optimising with respect to the full trajectories, we instead split the data into batches of horizon $H$.
The optimal parameters are found by minimising 
\[
\hat{\modelparam} = \min_{\modelparam} \c{L}_T + \c{L}_C + \c{L}_R
\]
using mini-batch stochastic gradient descent.

\paragraph{$\ell^2$ regularisation.}

Since the change in velocity in CD-Lagrange $\m{\dot Q}_{\npth} = \m{\dot Q}_{\npth}^S + \m{\dot Q}_{\npth}^C$ is \emph{additive} over conservative and contact components, as part of training, a CD-Lagrange network needs to learn to distinguish which component should be used to predict a given trajectory in the training data.
$\ell^2$ regularization affects this aspect of the learning problem in an subtle but outsized manner: as a consequence of shrinking the potential network's weights to zero, the regularizer shrinks the \emph{function} $\v{F}(\m{Q}_{n+1}): \R^d \-> \R^d$ to the \emph{zero function} $\bb{0} : \m{Q} \|> \v{0}$.
This encourages the network to explain contact events with contact forces rather than conservative forces where possible.

\section{Experiments}

In order to investigate the properties of the proposed CD-Lagrange networks, we run experiments on a number of reference rigid body systems.
In Section \ref{sec:ideal-pendulum}, we study performance on an ideal pendulum system without contacts, focusing on the physical properties of the conserved dynamics defined in equations \eqref{eq:cdl-smooth} and \eqref{eq:cdl-conservative}, thereby checking the network's performance relative to baselines.
In Section \ref{sec:bouncing-ball}, we perform experiments on a rigid bouncing ball system, exploring the behaviour of the network when contact happens and the effect of including an idealised touch sensor on the ability of recovering the underlying dynamics.
Finally, in Section \ref{sec:newton-cradle}, we study body-body impacts in an idealised Newton's cradle and look at the network's ability to recover the underlying dynamics and effects of numerical interpenetration during contact events.

We compare CD-Lagrange networks with residual networks (ResNets) and variational integrator networks (VINs) by evaluating them in terms of predictive performance on held-out test data as well as qualitatively evaluating the corresponding phase diagrams.
For contact experiments, we additionally include a modified residual network baseline (ResNetContact), which takes as input both the state as well the contact signal, for comparison against the CD-Lagrange networks' idealised touch feedback sensor.
To generate data, we simulate trajectories of motion and add independent Gaussian noise to the positions and velocities.
Full details for experiment hyperparameters and training are given in Appendix \ref{apdx:exp}.

\subsection{Learning to predict motion without contacts: an ideal pendulum}\label{sec:ideal-pendulum}

We first examine whether the performance of CD-Lagrange networks matches previous work on physically structured networks in cases where there is no contact. To this end, we consider learning to predict the trajectory of a simple \emph{ideal pendulum} from observed data.
The ideal pendulum is a point-mass attached to a mass-less rigid rod, suspended from a pivot. 
The pendulum swings back and forth due to gravity, without friction, so that the system conserves energy. 
This task has been studied previously by a number of authors, such as \textcite{greydanus2019hamilton} and \textcite{saemundsson2020variational}, who propose to use a network constructed from a variational \emph{velocity Verlet} integrator in order to learn in a data-efficient manner.
The variational integrator network preserves more physical properties than the CD-Lagrange network, and therefore we expect it to perform better in settings where there are no contacts.
We generate $20$ training trajectories consisting of $10$ points each by simulating trajectories and adding independent Gaussian noise to all position-velocity pairs.

\begin{figure}[b!]
    \centering
    \begin{tikzpicture}[scale=0.8]
        \centering
        \draw (-3, 0) -- (3,0);
        \draw [fill=white, thick] (0,2) circle (0.4);
        \draw [-latex, thick] (1,0) -- (1, 1) node[pos=0.5, right]{$\v{n}$};
        \draw [-latex, thick] (0,1.6) -- (0, 0.6) node[pos=0.5, left]{$-mg$};
        \draw [-latex, thick] (-2,0) -- (-2, 2) node[pos=0.5, left]{$\m{x}$};
        \fill[fill=black] (-3,0) -- (3,0) -- (3,-0.1) -- (-3,-0.1);
    \end{tikzpicture}
    \caption{Schematics of the bouncing ball. Here, the ball falls due to the gravitational pull $-mg$, and experiences a contact force in the direction of the contact normal vector $\v{n}$ with respect to the floor.}
    \label{fig:bouncing_ball}
\end{figure}
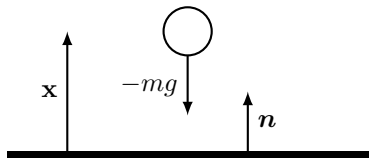

Figure \ref{fig:pendulum} plots the phase-space trajectories for the ground truth system, the observed training data and the predicted evolution for each model. 
The CD-Lagrange network and variational integrator network achieve comparable error, which significantly improves upon the baseline residual network, which incorrectly dissipates energy.
The network also recovers the correct contact function, which is the zero function.
This shows that relative to prior works on physically structured neural networks, accurate performance in contact-free scenarios is not compromised by the extra structure added to handle contacts.

\subsection{Learning body-wall contacts: an ideal bouncing ball}
\label{sec:bouncing-ball}

\begin{figure*}[t!]
    \centering    
    \includegraphics[scale=0.45]{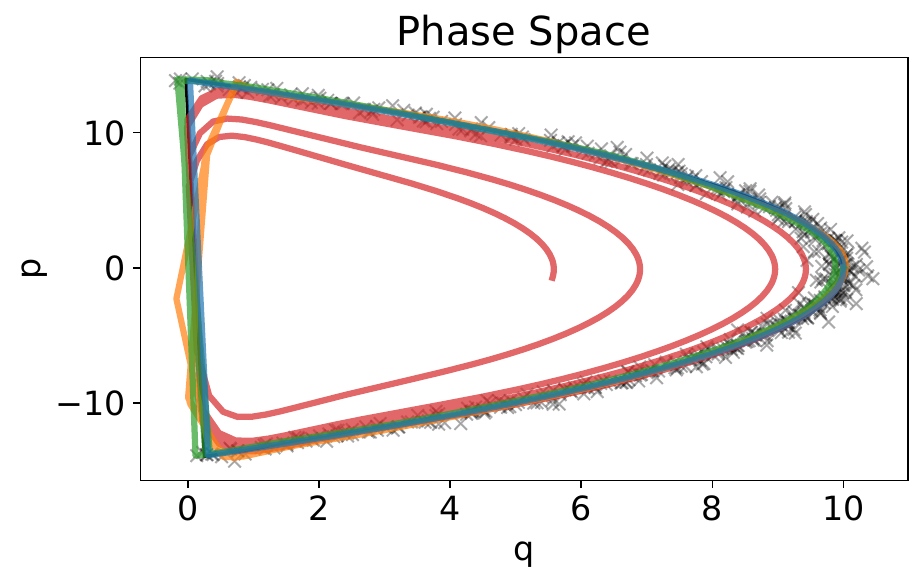}
    \includegraphics[scale=0.45]{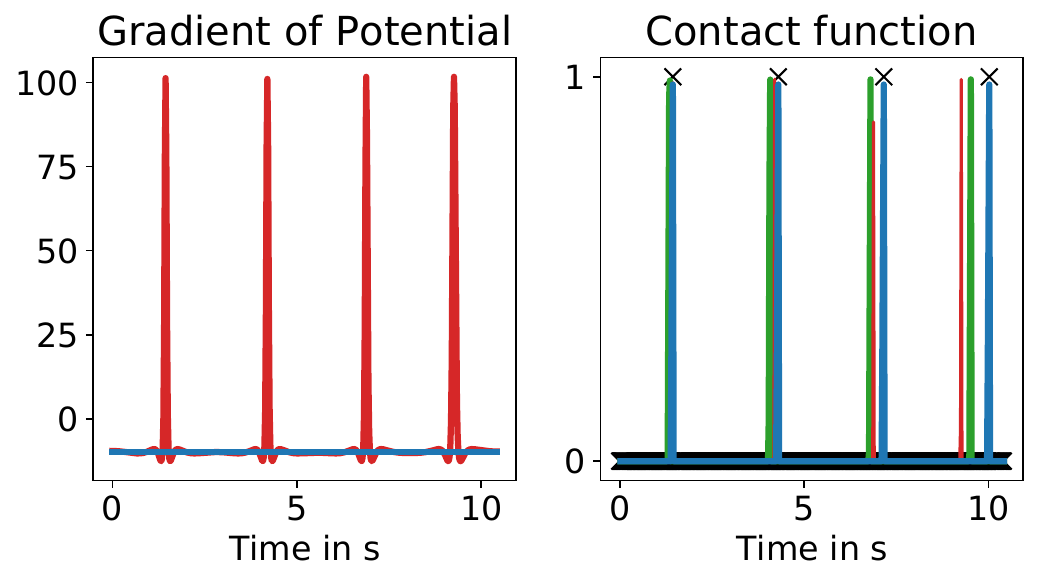}
    \includegraphics[scale=0.5]{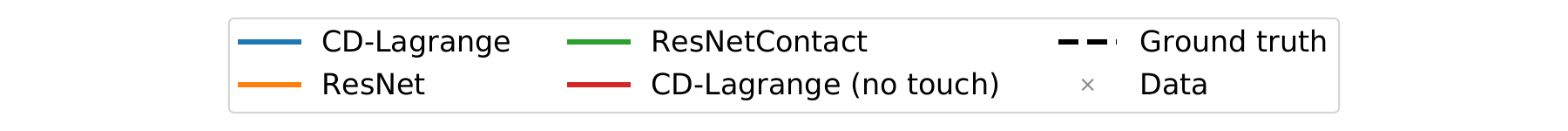}
    \caption{Learning the equations of motion of the bouncing ball. Given a set of initial conditions, we forecast a path in phase space and predict against ground truth. We display ground truth, training data, and model predictions for each of the models. For the CD-Lagrange network, we display the gradient of the potential energy, and the learned contact function.
    Root mean squared errors for each network's predictions are as follows: \textsc{CD-Lagrange: 0.067, CD-Lagrange (no touch): 7.578, ResNet: 6.778, ResNetContact: 8.866.}}
    \label{fig:bouncing-ball}
\end{figure*}

Next, we study using CD-Lagrange networks to learn contact dynamics for an ideal rigid \emph{bouncing ball}.
This system, shown in Figure \ref{fig:bouncing_ball}, is a commonly studied example in the non-smooth contact dynamics literature \cite{di2019benchmark}. 
The ball accelerates towards the ground due to the force of gravity, hits the ground, and bounces back up.
The effect of the impact is determined by Newton's restitution law, given in equation \eqref{eq:restitution-law}, where the elasticity parameter $e$ controls the energy behaviour. 
We focus on the regime $e=1$, where energy is conserved.

\begin{table}[b!]
\centering
\begin{tabular}{l ll}
              & RMSE \\
\midrule
ResNet        & $6.6\pm 1.2$  \\
ResNetContact & $4.8\pm 0.8$ \\
CD-Lagrange   & $1.9\pm 1.0$
\end{tabular}
\caption{Average root-mean squared error and standard error of the bouncing ball experiment averaged over $5$ runs with $40$ trajectories each consisting of $10$ data points.}
\label{tbl:ball-rmse}
\end{table}

We consider two different data regimes.
In the first regime, only observed trajectories of motion are provided at training time.
In the second regime, we add \emph{idealised touch sensor data} that indicates at each time point whether or not contact has occurred, given in equation \eqref{eq:touch-sensor}.
This can be viewed as representing an ideal impact sensor located inside the ball, or along the floor.
To learn this system's equations of motion from data, we generate $52$ training trajectories of $10$ points each, and train a CD-Lagrange network, a residual network, and a modified residual network that also takes contact data as input.
Results can be seen in Figure \ref{fig:bouncing-ball}.
Here, we see that the CD-Lagrange network's performance depends critically on whether or not it has access to touch feedback data.
With touch feedback, the CD-Lagrange network predicts the ball's trajectory much more accurately than the residual network. 
Without touch feedback, the CD-Lagrange network fails to learn the correct dynamics---instead, the network attempts to incorrectly explain noise using contact events, and contact events using smooth dynamics, because all scenarios lead to similar-looking noisy data from the network's perspective.
The residual network struggles to approximate the non-smooth behaviour at impact time, replacing instantaneous contacts with fast movement.
Adding contact information to the residual network's inputs does not improve its performance.

From examining the potential and contact network in Figure \ref{fig:bouncing-ball}, we see that the CD-Lagrange network with touch feedback determines the impact times near-exactly. 
The gradient of the potential energy remains very close to the ground truth value, even as the system evolves and contact events occur.
This shows that the network correctly determines that contact-driven changes in system states are caused by contacts, and not by spurious potential energy within the smooth dynamics---so long as the network is provided with touch feedback that enables it to differentiate between contact events and noise.
Root mean squared error, together with standard error, can be seen in Table \ref{tbl:ball-rmse}.

\subsection{Learning body-body impacts: Newton's cradle}
\label{sec:newton-cradle}

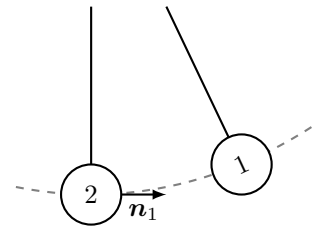
\begin{figure}[b!]
    \centering
    \begin{tikzpicture}
        \centering
        \draw [thick, domain=-1:3, dashed, gray] plot (\x, {\x*tan(asin(\x/10))});
        \draw[thick] (0, 0) -- (0,2.5);
        \draw [thick,fill=white] (0,0) circle (0.4) node[align=center] {2};
        \draw[thick] (2, 0.4) -- (1,2.5);
        \draw [thick,fill=white] (2,0.4) circle (0.4) node[align=center, rotate=25] {1};
        \draw [-latex,thick] (0.4,0) -- (1, 0) node[pos=0.5, below]{$\v{n}_1$}; 
    \end{tikzpicture}
    \caption{Schematics of the Newton's cradle system with two balls. Here, the first ball swings along the suspended rope due to gravity, and experiences a contact force in the direction of the contact normal vector $\v{n}_1$ upon impact.}
    \label{fig:newton_cradle_intro}
\end{figure}

Finally, we consider learning body-body impacts using CD-Lagrange networks in a simple \emph{Newton's cradle} system consisting of two balls suspended by a string, shown in  Figure \ref{fig:newton_cradle_intro}.
We assume both bodies have no volume, are suspended from a common mounting point, and parameterise their locations using angles relative to the vertical axis.
This means that collisions will occur perpendicular to the contact surface.
We train on $54$ trajectories consisting of $10$ data points each, and again consider CD-Lagrange networks with and without idealised touch sensor data, along with residual network baseline within both data regimes.

\begin{figure*}[t!]
    \centering
    \includegraphics[scale=0.45]{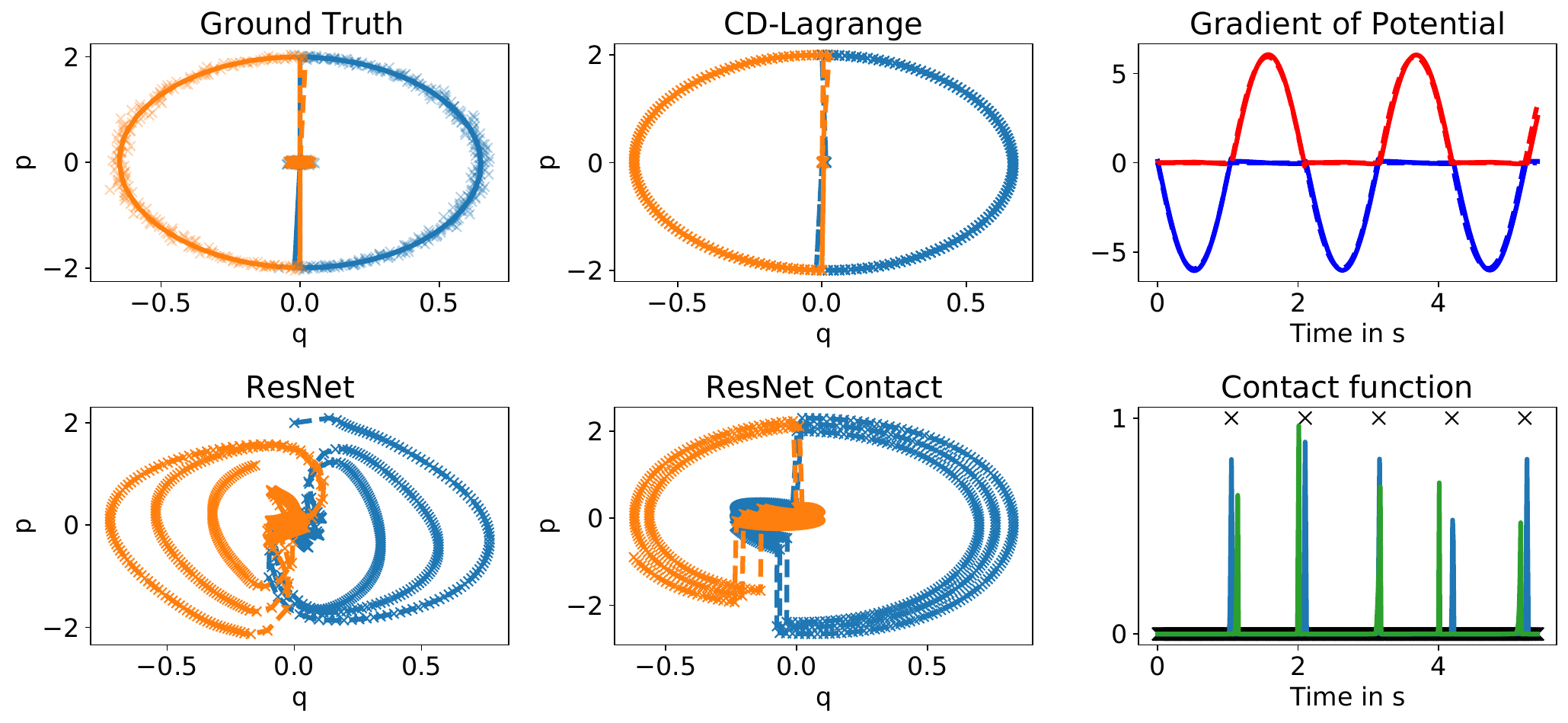}
    \includegraphics[scale=0.5]{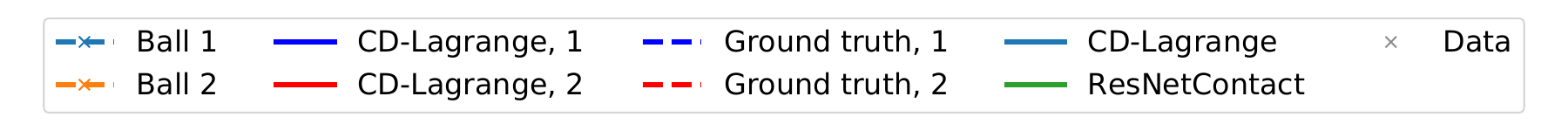}
    \caption{Learning the equations of motion of the Newton's cradle. Given a set of initial conditions, we forecast a path in phase space and predict against ground truth. We display ground truth, training data, and model predictions for each of the models. For the CD-Lagrange network, we display the gradient of the potential energy, and the learned contact function. Root mean squared errors for each network are given as follows: \textsc{CDL RMSE: 0.406, Resnet RMSE: 1.685, ResnetContact RMSE: 1.034.}}
    \label{fig:newton-cradle}
\end{figure*}

\begin{table}[b!]
\centering
\begin{tabular}{l ll}
              & RMSE            \\
\midrule
ResNet        & $1.6\pm 0.1$   \\
ResNetContact & $3.5\pm 1.3$  \\
CD-Lagrange   & $0.4\pm 0.1$ 
\end{tabular}
\caption{Average root-mean squared error  with standard error of the Newton's cradle experiment averaged over $5$ runs with $50$ trajectories each consisting of $10$ data points.}
\label{tbl:nc-rmse}
\end{table}

Results can be seen in Figure \ref{fig:newton-cradle}.
Here, we see that the phase-space trajectory of the CD-Lagrange network is substantially more accurate than for a baseline residual network.
Root mean squared error, together with standard error, can be seen in Table \ref{tbl:nc-rmse}.
As before, adding contact information does not improve the residual network's performance. 
This replicates the results of the single-body bouncing ball experiment in Section \ref{sec:bouncing-ball} in a multi-body setting.

Next, we examine how accurately the potential energy and contact function are recovered---this can also be seen in Figure \ref{fig:newton-cradle}.
The CD-Lagrange network learns the potential accurately, with some error around the contact points.
In particular, the contact network accurately learns to determine when contact will occur based on the touch sensor data.
Mirroring the results of Section \ref{sec:bouncing-ball}, the residual network struggles to predict contact events even when it is given explicit touch sensor data on when they occur, particularly through mistiming contact events to occur earlier or later than is correct.

The CD-Lagrange integrator, and by proxy the CD-Lagrange network, allow some slight interpenetration to occur.
In body-wall impact scenarios, this does not strongly affect performance. 
However, in body-body impacts, this qualitative behaviour difference becomes more significant.
Specifically, as momentum gets transferred between bodies, the slight allowed interpenetration can cause the system to significantly violate conservation of energy. 
In the Newton's cradle example, this causes the resting ball to gain potential energy during collision events, eventually causing both balls to accelerate.
To mitigate this problem, we introduce a closest-point projection during impact events \cite{fetecau2003nonsmooth}, which aligns the resting ball with the closest point on the boundary where no penetration occurs.
This improves accuracy and restores correct long-term system behaviour.

\section{Discussion}

CD-Lagrange networks are a flexible way for data-driven learning of equations of motion that include contact dynamics.
These networks can learn to accurately predict trajectories and resolve collisions in a noise-robust and data-efficient manner.
This contributes to a rapidly-expanding line of work on neural ODEs and physically structured learning, and shows that these ideas can work successfully in non-smooth settings.

Compared to black-box neural networks, the implementation of physically structured networks for learning contact dynamics presents a number of additional challenges to ensure their performance.
In particular, we find that the presence or absence of touch feedback impacts the model's performance dramatically.
This is because without touch feedback the learning problem is ambiguous: abrupt shifts in the system state can be explained either by contacts, or by noise, and it is difficult for the network to tell which is which.
This results in weak identifiability in the loss function, which could lead to difficulties with local optima or other issues.
It is therefore important that the network is provided with an appropriate means to differentiate between the two, e.g. by means of a touch sensor.

The networks studied here are based on the CD-Lagrange integrator, which is an explicit scheme that fits conveniently within an automatic differentiation framework.
This convenience comes at a cost: the integrator allows for some physically incorrect interpenetration, which can affect long-term prediction accuracy.
Other schemes, such as variational integrators for contact dynamics \cite{fetecau2003nonsmooth}, avoid these issues and achieve higher accuracy, but they require the solution of fixed-point iterations or convex optimisation problems during time-stepping, which renders them more expensive and cumbersome to work with.
These issues can potentially be mitigated through the use of differential physics engines as components of deep-learning-based models \cite{de2018end}.
Studying these trade-offs in the context of learning could pave the way toward better understanding on how to incorporate inductive biases to improve performance of neural networks to model the real world, thereby facilitating their use in applications such as robotics.

\section{Conclusion}

In this work, we introduce CD-Lagrange networks, which build on ideas from neural ODEs and physically structured learning to construct networks for learning contact dynamics in data-limited regimes.
With the addition of an idealised touch feedback sensor, these networks can learn to accurately reconstruct non-smooth contact dynamics from data, in a way that disentangles contact-driven forces from conservative forces.
The simple and explicit structure makes these networks interpretable, well-matched with the underlying physics, and straightforward to implement.
A rapidly growing line of work has focused on adapting deep networks to various physical settings. We hope our contributions facilitate the inclusion of non-smooth contact dynamics within these settings.

\printbibliography

\onecolumn
\appendix

\section{Appendix: extended CD-Lagrange networks}
\label{apdx:cdl}

The CD-Lagrange scheme, in full generality, is designed for deformable body impacts for finite element calculations \cite{fekak2017new}. 
In the non-deformable case, \textcite{fekak2017new} propose employing Newton's restitution law for rigid body impacts. 
This variant has been benchmarked by \textcite{di2019benchmark}. We extend the framework to support three-dimensional rigid body-body collisions by ensuring that the overall momentum is conserved during the impact. 
Instead of multi-node bodies---as used, for example, in finite element simulations of deformation during multi-body collision---we restrict ourselves to single-node bodies in order to simplify the problem.

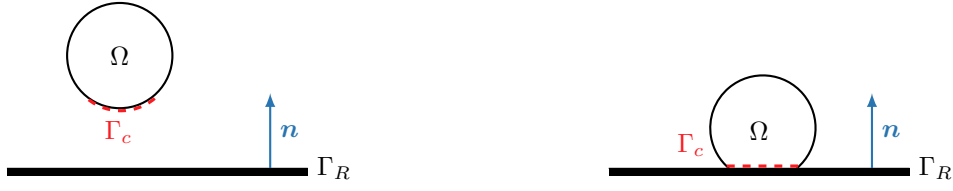
\begin{figure}[t!]
    \centering
    \begin{tikzpicture}
        \draw (-6, 0) -- (-2,0) node[right]{$\Gamma_R$};
        \draw [thick, fill=white] (-4.5,1.5) circle [x radius=0.7, y radius=0.7, fill=white] node[align=center] {$\Omega$};
        \draw [-latex, plotblue, thick] (-2.5,0) -- (-2.5, 1) node[pos=0.5, right]{$\v{n}$};
        \fill[fill=black] (-6,0) -- (-2,0) -- (-2,-0.1) -- (-6,-0.1);
        \draw [dashed,plotred,very thick] (-4.035,0.93) arc [x radius=0.7, y radius=0.7, start angle=-50, end angle=-135] node[pos=0.5, below]{$\Gamma_c$};
        
        \draw [thick, fill=white] (4.5,0) arc [x radius=0.7, y radius=0.7, fill=white, start angle=-50, end angle=230];
        \draw (4.0, 0.5) node[align=center] {$\Omega$};
        \draw (2, 0) -- (6,0) node[right]{$\Gamma_R$};
        \draw [-latex, plotblue, thick] (5.5,0) -- (5.5, 1) node[pos=0.5, right]{$\v{n}$};
        \fill[fill=black] (2,0) -- (6,0) -- (6,-0.1) -- (2,-0.1);
        \draw [dashed,plotred,very thick] (3.57,0.03) -- (4.52,0.03) node[pos=-0.5, above]{$\Gamma_c$};
    \end{tikzpicture}
    \caption{Visualising a deformable body $\Omega$ before (left) and during (right) impact into a rigid wall $\Gamma_R$. Here, $\v{n}$ denotes the surface normal of the wall and $\Gamma_c$ the set of contact points.}
    \label{fig:cdl-boundary}
\end{figure}

CD-Lagrange is based on an asynchronous version of the central difference method and uses Lagrange multipliers $\lambda^{(k)}$ to enforce the contact constraints.
For a given problem, $\v{\lambda}$ must fulfill the Karush--Kuhn--Tucker (KKT) conditions, sometimes also referred to as Hertz--Signorini--Moreau (HSM) conditions \cite{wriggers1999new}.
These are given by
\[
g_N = (\v{x}_M - \v{x})^T \v{n} \geq 0
\label{eq:HSM:gN}
\\
\v{\lambda}_N = -\v{n}_\Omega \cdot \v{\sigma}(\v{x}) \cdot \v{n}_\Omega \geq 0
\label{eq:HSM:lambda}
\\
g_N(\v{x}) \v{\lambda}_N = 0
\label{eq:HSM:gNlambda}
\]
which hold for all $\v{x} \in \Gamma_c$, where $\v{n}$ is the outer pointing normal on the wall $\Gamma_R$ and $\v{n}_\Omega$ the outer-pointing normal on the deformable body $\Omega$. 
In particular,  $\v{n}_\Omega = -\v{n}$ when a contact occurs. 
$\Gamma_c$ represents all contact points on $\Omega$, $\v{x}_M \in \Gamma_R$ the closest point to $\v{x} \in \Gamma_c$ and $\v{\sigma}$ is the Cauchy stress field.

Condition \eqref{eq:HSM:gN} enforces the impenetrability, \eqref{eq:HSM:lambda} implies that no adhesion occurs upon contact which means the contact-stress is only compressive, and \eqref{eq:HSM:gNlambda} reduces the contact-states to \textsc{contact} ($g_N = 0$ and $\lambda_N > 0$) and \textsc{no contact} ($g_N > 0$, $\lambda_N=0$). The subscript $N$ is to emphasise dependence on the \emph{normal} part of the velocity, i.e. the part perpendicular to the contact surface. This subscript will be omitted in the rest of this derivation to simplify notation.

We represent the position of all (single-node) bodies in a $d$-dimensional space by
\[
\m{Q} = \left(\v{q}^k, k \in \{1,\ldots, K\}\right)
\]
where $\m{q}^k \in \mathbb{R}^d$ represents the individual position of body $k$.

To be able to calculate the velocity component each body has towards the impact surface a projection operator
\[
\m{L} = \left((\v{n}^k), k \in \{1, \ldots, K\} \right)
\]
is introduced, where $\v n^k \in \mathbb R ^{d}$ is the surface normal of the object body $k$ is colliding with.
The distance between each body and the closest point on the wall or another body is represented by
\[
\v{g} = \left(g^k, k \in \{1,\ldots,K\}\right), \quad \text{where}\quad g^k(\m Q) \in \mathbb{R}.
\]
Finally, define $\v{c} = \sgn(\v{g})$ to be the \emph{contact function}. Note that the equations of motion depend only on $\v{g}$ through $\v{c}$: we will subsequently replace this function with the contact network $\v{\hat c}_\theta \in \{0, 1\}$.
\paragraph{Discrete impact equations.}
Given a single impact event at time $t_c$ of a hyperelastic solid, we may write the action integral \cite{cirak2005decomposition} 
\[
S(\m{Q}, \m{\dot Q}, t_c) = \int_0^{t_c} L(\m{Q}, \m{\dot Q})\mathrm{d}t + \int_{t_c}^T L(\m{Q}, \m{\dot Q})\text{d}t
\]
where $L(\m{Q}, \m{\dot Q})$ is the semi-discrete Lagrangian of the form
\[
L(\m{Q}, \m{\dot Q}) = \m{\dot Q}^T \m{M} \m{\dot Q} + V(\m{Q})\text{,}
\]
where $\m{M}$ represents the mass matrix and $V$ the external potential, e.g. the gravitational field \cite{cirak2005decomposition}.
If there are multiple impacts over the interval $[0, T]$ the action integral can be split up further to handle the contact events in a sequential manner.
We apply small variation of the path $\delta q$ 
\[
    \delta S(\m{Q}, \m{\dot Q}, t_c) &= \delta \left( \int_0^{t_c} L(\m{Q}, \m{\dot Q})\mathrm{d}t + \int_{t_c}^T L(\m{Q}, \m{\dot Q}, t_c)\mathrm{d}t \right)\\
    &= \int_0^T \left( \dfrac{\partial L}{\partial \m{Q}} - \dfrac{\mathrm{d}}{\mathrm{d}t} \dfrac{\partial L}{\partial \m{\dot Q}}\right)\delta \m{Q} \text{d}t - \left[ \dfrac{\partial L}{\partial \m{\dot Q}} \cdot \delta \m{Q} + L \delta t\right]_{t_c^-}^{t_c^+} = 0
    .
\]
Applying the Fundamental Lemma of the Calculus of Variations \cite{cirak2005decomposition} yields the equations
\[
    \m{\ddot Q} = \m{M}^{-1} \dfrac{\partial V(\m{Q})}{\partial \m{Q}} \label{eq:equilibrium}
    \\
    \left[ \m{M \dot Q} \right]_{t_c^-}^{t_c^+} = \m I (\v{\lambda}(t_c)) \label{eq:non-smooth1}
    \\
    \left[ \left(\m{M \dot Q}\right)^T \m{M^{-1}} \left(\m{M \dot Q}\right) \right]_{t_c^-}^{t_c^+} &= 0 \label{eq:non-smooth2}
\]
which fully describe the dynamics. 
Here \eqref{eq:equilibrium} represents the smooth dynamics, \eqref{eq:non-smooth1} represents the non-smooth dynamics and \eqref{eq:non-smooth2} represents the kinetic energy balance upon impact.

Using the theory of non-smooth dynamics \cite{moreau1988unilateral}, \eqref{eq:equilibrium}, \eqref{eq:non-smooth1}, and \eqref{eq:non-smooth2} can be combined into one equation and the velocity can be expressed as a sum of the smooth and non-smooth parts, given by
\[
\mathrm{d} \m{\dot Q} = \mathrm{d}\m{\dot Q}_\text{s} + \mathrm{d} \m{\dot Q}_\text{ns}.
\]
\textcite{fekak2017new} show that incorporating the KKT conditions into the above yields the \emph{non-smooth contact dynamics} (NSCD) equations
\begin{gather}
\m{M}\mathrm{d}\m{\dot Q} = \dfrac{\partial V(\m{Q})}{\partial \m{Q}} + \mathrm{d}\m I(\v \lambda)
\\
    \forall k \in \{ 1, \ldots, K \} \quad \begin{cases}
    0 \leq \lambda^k \perp v^k \geq 0\text{,} & \text{if } g^k = 0\\
    \lambda^k = 0\text{,} & \text{if } g^k > 0.
    \end{cases} \label{eq:nscd}
\end{gather}
By integrating \eqref{eq:nscd} between $t_{n+\frac{1}{2}}$ and $t_{n + \frac{3}{2}}$ and estimating the time integrals using a midpoint-rule, following \textcite{fekak2017new}, one obtains the semi-discrete equilibrium equation
\[
    \m{M}\left(\m{\dot Q}_{n+\frac{3}{2}} - \m{\dot Q}_{n+\frac{1}{2}} \right) = h \dfrac{\partial V(\m{Q})}{\partial \m{Q}} + \m{I}(\v{\lambda}_{n + \frac{3}{2}})\text{,} \label{eq:discrete-equilibrium}
\]
which results in an asynchronous update rule for the velocity given by
\[
   \m{\dot Q}_{n+\frac{3}{2}} &= \m{\dot Q}_{n+\frac{3}{2}}^S + \m{M}^{-1} \m I(\v \lambda_{\npth}),
   \label{eq:CDL:general_impulse}
   &
    \m{\dot Q}_{n+\frac{3}{2}}^S &= \m{\dot Q}_{n+\frac{1}{2}} + \m{M}^{-1} h  \dfrac{\partial V(\m{Q})}{\partial \m{Q}}.
\]
Further, $\m{Q}$ can be updated using the previous velocity
\[
\m{Q}_{n+1} = \m{Q}_{n} + \dfrac{h}{2} \m{\dot Q}_{n+\frac{1}{2}}
\]

To calculate the impulse $\m I$ we combine Newton's restitution law, as proposed by \cite{fekak2017new}, and the law of conservation of linear momentum. 
Consequently, the impulse between body $i$ and $j$ caused by a collision is defined by
\[\label{eq:two-body-momentum-transfer}
\v{I}_{ij} = (1 + e) \dfrac{m_i m_j}{m_i + m_j} \v n_i \left( \v{\dot q}^j - \v{\dot q}^i \right) \v{n}_i.
\]
Unpacking this equation, the velocity difference between the two bodies is projected onto the normal of the contact surface $n$ and scaled by the mass ratio $m_i m_j/ (m_i + m_j)$ and the elasticity $e$.

We introduce the matrix $\m A$ which, at impact times, determines which bodies are in contact.
This is given by
\[
[\m A_n]_{ij} =
\begin{cases}
-1 & \text{if  $i = j$ and } [\v g_n]_i < 0\\
1  & \text{if body $i$ and $j$ are in contact}\\
   & \text{(implicitly $[\v g_n]_i = [\v g_n]_j < 0$)}\\
0  & \text{otherwise.}
\end{cases},\quad \sum_{i=1}^K \sum_{j=1}^K [\m A_n]_{ij} = 0
.
\]
From this, the mass ratio can be expressed as
\[
[\m H]_{ij} = [\m A \m M ^{-1} \m A^T]^{-1}_{ij}
\]
where the inverse is applied element-wise.
This definition of $\m A$ works for multiple simultaneous two-body collisions. 
In case three (or more) bodies collide with each other simultaneously, contacts are resolved sequentially.

Solving equation \eqref{eq:discrete-equilibrium} for $\m I$ and using equation \eqref{eq:two-body-momentum-transfer} to allow two-body impacts yields
\[
\v I_{n+1}^k &= \v L^k_{n+1} \max \left(0, \lambda^k_{n+1} \right)
&
\lambda^k_{\npth} &= \left[\m H_{n+1} (e \m{\dot Q}_{\nph} + \m{\dot Q}^S_{\npth})\m L_{n+1}^T \right]_{kk}.\label{eq:r_k}
\]
where $\v{L}^k = \v{n}_k$.
In equation \eqref{eq:r_k}, the operator $\m H$ incorporates the mass ratio, and is responsible for selecting the two bodies that are interacting in the collision when applied to $\m{\dot Q}$. $\m L$ projects the velocity for every body onto the surface normal of the contact surface. 
Since we work with independent single-node bodies, we are only interested in the diagonal elements of $\m{H}$.
The maximum makes sure the velocity of the colliding object is not pointing away from the surface, indicating that the collision already happened in the previous time step.
We replace $\v g(\m Q)$ with the contact network $\v{\hat c}_\theta(\m Q, \m{\dot Q})$.
Since the contact network $\v{\hat c}_\theta$ receives $\m Q$ as well as $\m{\dot Q}$, it can learn whether the collision already happened or not, and thus we omit the $\max$.

Summarising, the integrator is governed by the equations
\[
    &\m{Q}_{n+1} = \m{Q}_n + h \m{\dot Q}_{n+\frac{1}{2}}
    ,\label{eq:sum_u}\\
    &\m{L}_{n+1} = \m{L}(\m Q_{n+1}), \quad \v{\hat{c}}_{n+1} = \v{\hat c}_\theta(\m{Q}_{n+1}, \m{\dot Q}_{\nph}), \quad \m{H}_{n+1} = \m{H}(\m{Q}_{n+1}, \m{\dot Q}_{\nph})\\
    & \m{F}_{n+1} = \dfrac{\partial V(\m{Q}_{n+1})}{\partial \m{Q}_{n+1}}
    ,\label{eq:sum_F} \\
    &\m{U}_{n+\frac{3}{2}} = - e \m{\dot Q}_{n + \frac{1}{2}} \m{L}_{n+1}^T
    ,\\
    &\begin{cases}
        \text{if } \v{\hat{c}}^k_{n+1} = 1 & \v{I}^k_{n+\frac{3}{2}} = \v L^k_{n+1} \left[ \m{H} \left( \m{U}_{n+\frac{3}{2}} + \left( \m{\dot Q}_{n+\frac{1}{2}} + h \m{M}^{-1} \m{F}_{n+1} \right)\m L^T_{n+1} \right)\right]_{kk}\\
        \text{otherwise} & \v{I}^k_{n+\frac{3}{2}} = \v 0
    \end{cases}
    ,\label{eq:rN}\\
    &\m{\dot Q}_{n+\frac{3}{2}} = \m{\dot Q}_{n+\frac{1}{2}} + \m{M}^{-1}\left( h \m{F}_{n+1} + \m I_{n+1} \right)\label{eq:sum_udot}
    .
\]

In our experiments we assume $\m H$ is known. $\v{\hat c}_\theta$ represents the contact-network, which returns $1$ if a contact occurs and $0$ if not. The potential $\v V$ is approximated using the potential-network.

We conclude with two remarks on more general settings.
Firsly, if we wish to allow multiple nodes per body the operator $\m{H}$ needs to be adapted accordingly, and the internal stress field $\v{W}$ has to be added to the force (\ref{eq:sum_F}), to yield
\[
\m{F}_{n+1} = \dfrac{\partial V(\m{Q}_{n+1})}{\partial \m{Q}_{n+1}} - \dfrac{\partial W(\m{Q}_{n+1})}{\partial \m{Q}_{n+1}}
.
\]
Further, it is no longer possible to use only the diagonal elements of the impulse equation \eqref{eq:rN}, as now the interactions between nodes play a role in contact resolution.

Secondly, the force $\m F$ can easily be extended to contain other forces. For example, we can add damping to obtain
\[
\m{F}_{n+1} = \dfrac{\partial V(\m{Q}_{n+1})}{\partial \m{Q}} - \m{C \dot Q},
\]
where $\m{C}$ is a learned damping coefficient.

\section{Appendix: experimental details}
\label{apdx:exp}

\paragraph{Learning rate.} We use a learning rate of $10^{-3}$ in combination with the ADAM optimiser in all experiments.

\paragraph{Potential.} We use the set-up proposed by \textcite{saemundsson2020variational}, which consists of a two-layer fully connected neural network with $\tanh$ activation function between the layers. 
The input layer has a size of 500 units and outputs a single real scalar. $\ell^2$ regularisation is used in both layers.

\paragraph{Baseline contact-aware residual network.}

To ensure a fair comparison, we use a vanilla residual network (ResNet) as a baseline. It consists of two dense layers with $\tanh$ activation function and an input layer size of $500$ units. We update the state according to
\[
\v s_{n+1} = \f{ResNet}(\v s_n, \v c_n).
\]
The residual network also receives the contact information $c_n$ and uses the same architecture as $\hat{c}_\theta$ to predict contact by means of
\[
\v c_{n+1} = \hat{c}_\theta^{\f{ResNet}}(\v s_n)
.
\]

\paragraph{Pendulum.} The parameters used in the pendulum experiment are
\[
g&=9.81\,\frac{\text{m}}{\text{s}^2} 
&
q_0 &= 1 
&
M &= 1\,\text{kg} 
&
h &= 0.02\,\text{s} 
&
l &= 1\,\text{m.}
\]
We add centered Gaussian noise with standard deviation $\sigma=0.2$ to the positions and velocities.
All models are trained for $3000$ epochs.

\paragraph{Bouncing ball.}

The projection operator is $L = 1$, and the other parameters are
\[
    e &= 1 
    &
    g&=9.81\,\frac{\text{m}}{\text{s}^2} 
    &
    q_0 &= 10\,\text{m}
    &
    M &= 1\,\text{kg}
    &
    h &= 0.02\,\text{s} 
    &
    \sigma &= 0.2
    .
\]
All models are trained for $2000$ epochs.

\paragraph{Newton's cradle.}
We will use two generalised angles to determine the position of both balls. The initial positions and velocities are
\[
\m{Q}_0 &= \begin{pmatrix}0\\0
\end{pmatrix}\text{,}
&
\m{\dot Q}_0 &= \begin{pmatrix}2\\0
\end{pmatrix}
\]
where $(\m{Q}^{(1)}_0, \m{\dot Q}^{(1)}_0)$ represent body 1, and $(\m{Q}^{(2)}_0, \m{\dot Q}^{(2)}_0)$ represent body 2.
The restitution parameter is $e=1$, and
\[
\m{A} = \begin{pmatrix} -1 & 1\\
1 & -1
\end{pmatrix}\text{.}
\]
For simplification, we assume both bodies have no volume and are connected with the same joint. 
This means collisions will always happen perpendicular to the contact surface. 
The projection operator is thus
\[
\m{L} = \begin{pmatrix} 1\\-1
\end{pmatrix}\text{.}
\]

\begin{figure}
    \centering
    \includegraphics[scale=0.45]{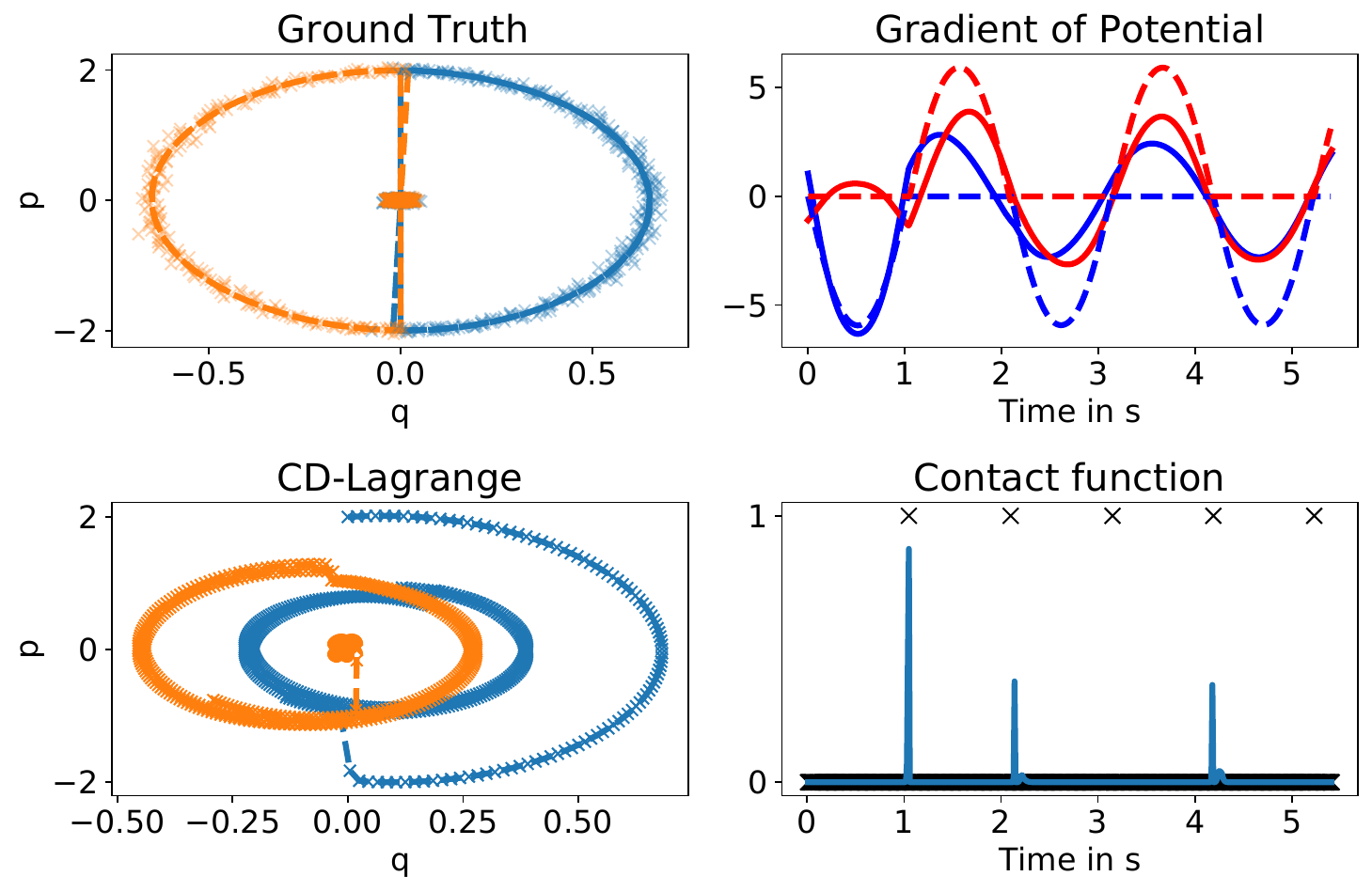}
    \includegraphics[scale=0.5]{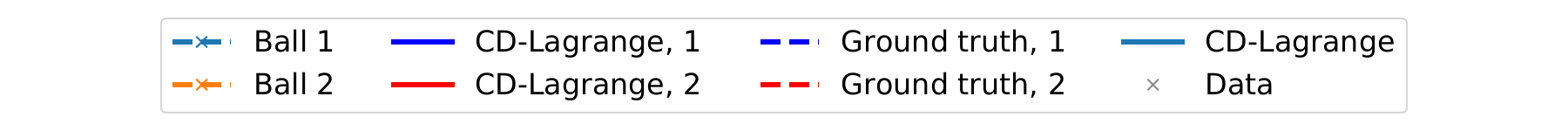}
    \caption{Learning the equations of motion of the Newton's cradle \emph{without touch feedback} and Gaussian noise in the training data with $\sigma = 0.02$. The model tries to approximate the collisions using the potential network and therefore struggles to predict the trajectory accurately. \textsc{CD-Lagrange RMSE: 0.907.}}
    \label{fig:newton-cradle-no-touch}
\end{figure}

\begin{figure}
    \centering
    \includegraphics[scale=0.45]{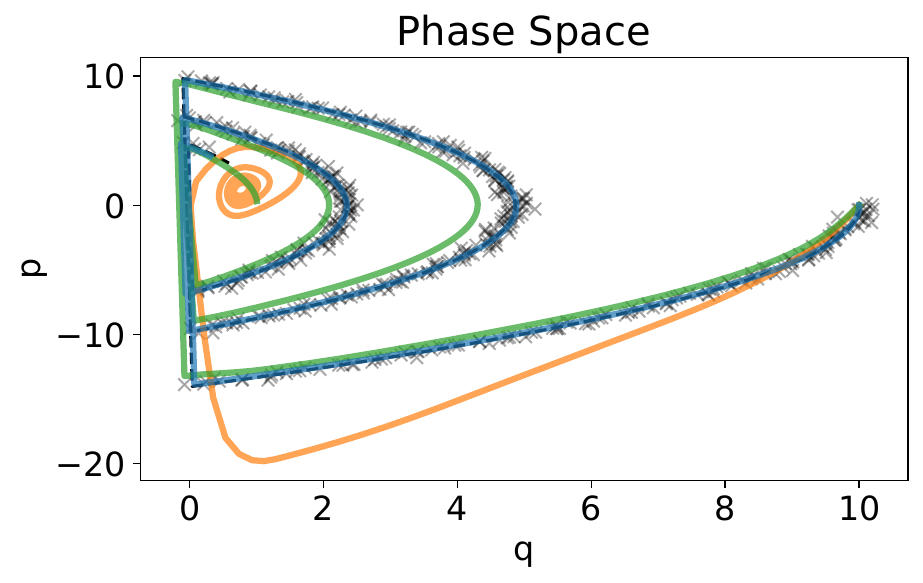}
    \includegraphics[scale=0.45]{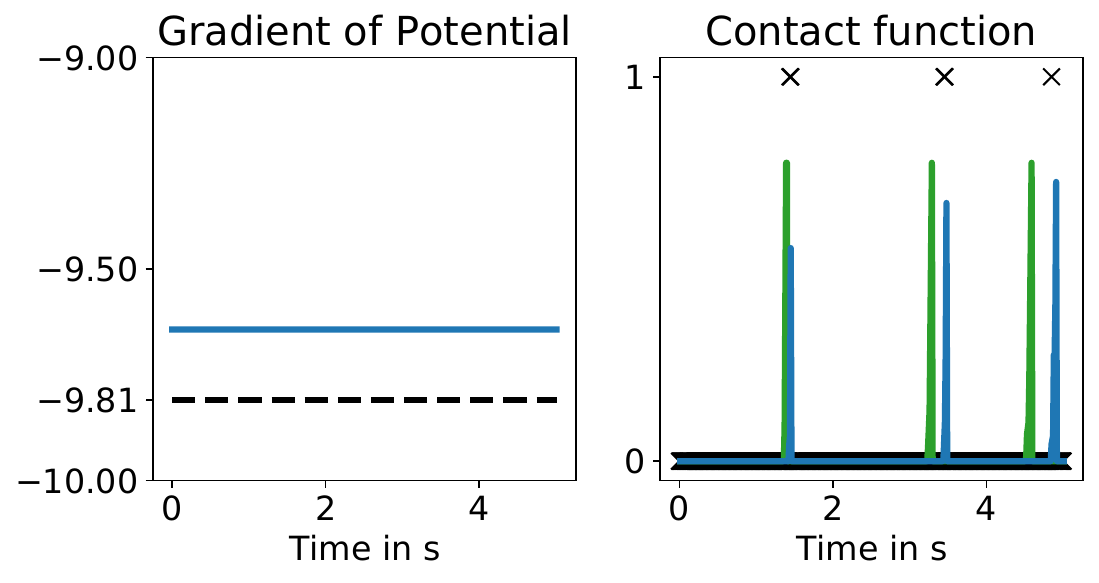}
    \includegraphics[scale=0.5]{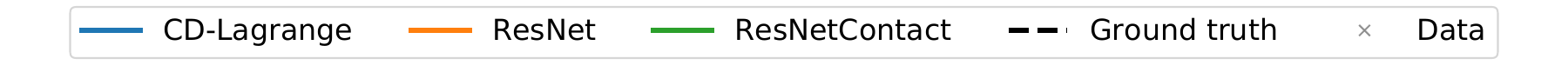}
    \caption{Learning the equations of motion of the bouncing ball with elasticity $e=0.7$. CD-Lagrange approximates the true trajectory in the most accurate manner, particularly for longer simulation times which correspond to phase space regions closer to the centre of the spiral. \textsc{CDL RMSE:2.076, ResNet RMSE: 8.291, ResNetContact RMSE: 4.156.}}
    \label{fig:bouncing-ball-elastic}
\end{figure}

\paragraph{Additional experiments.} Here we include additional experimental results for the Newton's cradle without touch feedback in Figure \ref{fig:newton-cradle-no-touch}, and the bouncing ball with elasticity $e = 0.8$ in Figure \ref{fig:bouncing-ball-elastic}.

\end{document}